\documentclass[lettersize,journal]{IEEEtran}
\usepackage{amsmath,amsfonts}
\usepackage{algorithmic}
\usepackage{algorithm}
\usepackage{array}
\usepackage[caption=false,font=footnotesize]{subfig}
\usepackage{textcomp}
\usepackage{stfloats}
\usepackage{hyperref}
\usepackage{verbatim}
\usepackage{graphicx}
\usepackage{cite}
\usepackage{paralist}
\usepackage{booktabs}
\usepackage{colortbl}
\usepackage{multirow}
\usepackage{balance}
\usepackage{tabularx}
\usepackage{lipsum}
\usepackage[table]{xcolor}

\newcommand{\avgstd}[2]{#1$\,\pm\,$#2}
\newcommand{\algcomment}[1]{\textcolor{gray!80}{\textit{#1}}}

\newcommand{\MEL}{MelSpec}
\newcommand{\HUB}{HuBERT}
\newcommand{\WAV}{Wav2Vec2}
\newcommand{\BERT}{BERT}
\newcommand{\ITB}{ItalianBERT}
\newcommand{\ROB}{XLMRoBERTa}

\newcommand{\finalDimension}{1280}
\newcommand{\lstmHiddenDimTimesTwo}{512}
\newcommand{\lstmHiddenDim}{256}
\newcommand{\attnHiddenDim}{128}
\newcommand{\crossAttnHiddenDim}{256}
\newcommand{\nHeadsCrossAttn}{4}
\newcommand{\fcHiddenDim}{128}

\newcommand{\weightDecay}{$1{\times}10^{-5}$}
\newcommand{\schedulerFactor}{0.5}
\newcommand{\schedulerPatience}{1}
\newcommand{\stopperEpsilon}{0.1}
\newcommand{\stopperPatience}{3}
\newcommand{\lambdaGRL}{0.5}
\newcommand{\numEpochs}{100}

\newcommand{\dropout}{0.1}
\newcommand{\nMels}{128}
\newcommand{\nFFT}{4096}
\newcommand{\hopLength}{512}

\newcommand{\impAcc}{2.5}
\newcommand{\impFscore}{3.3}
\newcommand{\accDG}{93.2}
\newcommand{\accDGstd}{5.7}
\newcommand{\precDG}{93.2}
\newcommand{\precDGstd}{9.7}
\newcommand{\recDG}{96.2}
\newcommand{\recDGstd}{4.7}
\newcommand{\FscoreDG}{94.2}
\newcommand{\FscoreDGstd}{4.7}

\makeatletter
\def\ps@IEEEtitlepagestyle{%
  \def\@oddhead{\parbox{\textwidth}{\centering \scriptsize This article has been accepted for publication in IEEE Transactions on Neural Networks and Learning Systems. This is the author's version which has not been fully edited and content may change prior to final publication. Citation information: DOI 10.1109/TNNLS.2026.3714047}}%
  \def\@evenhead{\@oddhead}%
  
  \def\@oddfoot{\parbox{\textwidth}{\centering \scriptsize \copyright 2026 IEEE. Personal use of this material is permitted. Permission from IEEE must be obtained for all other uses, in any current or future media, including reprinting/republishing this material for advertising or promotional purposes, creating new collective works, for resale or redistribution to servers or lists, or reuse of any copyrighted component of this work in other works.}}%
  \def\@evenfoot{\@oddfoot}%
}
\makeatother

\begin{document}

    \title{Multimodal Domain Generalization for Depression Detection: An Attention-Based BiLSTM Network with Domain-Adversarial Training}

    \author{Ali Tabaraei,
    Federico Simonetta, and
    Stavros Ntalampiras
    \thanks{Ali Tabaraei and Stavros Ntalampiras are with the Department of Computer Science, University of Milan, Milan, Italy. Emails: ali.tabaraei@unimi.it and stavros.ntalampiras@unimi.it}
    \thanks{Federico Simonetta is with the Computer Science Department, Gran Sasso Science Institute (GSSI), L’Aquila, Italy. Email: federico.simonetta@gssi.it}
    \thanks{Manuscript received 6 October 2025; revised 5 March 2026 and 13 June 2026; accepted 12 July 2026. (Corresponding author: Ali Tabaraei.)}}    
    \markboth{Journal of \LaTeX\ Class Files,~Vol.~14, No.~8, August~2021}%
    {Shell \MakeLowercase{\textit{et al.}}: Domain Generalization for Multimodal Depression Detection: An Attention-Based BiLSTM Network with Domain-Adversarial Training}
    
    \IEEEpubid{0000--0000/00\$00.00~\copyright~2021 IEEE}
    
    \maketitle

\begin{abstract}
    Automatic depression detection with deep learning has shown promise but often suffers from limited generalization due to domain shift arising from inter-speaker variability. To address this critical issue, we present the first patient-independent multimodal depression detection framework\footnote{Source code and documentation for this work can be accessed publicly at: \href{https://github.com/tabaraei/MultimodalDG-depression-detection}{https://github.com/tabaraei/MultimodalDG-depression-detection}} that incorporates domain generalization (DG), jointly leveraging both acoustic and textual modalities. The proposed model integrates \textit{bidirectional Long Short-Term Memory (BiLSTM)} with intra- and cross-modal attention mechanisms, accompanied by segment-level fusion for decision-making. Generalization is further enhanced by applying a \textit{gradient reversal} layer inspired by \textit{Domain-Adversarial Training of Neural Networks (DANN)}, which promotes domain-invariant representations by adversarially limiting the model's ability to identify individual speakers, effectively reducing patient-specific bias. Conducting experiments on the \textit{Androids-Corpus} dataset with a 5-fold cross-validation (CV) protocol, various pairings of audio and text feature extractors were evaluated over different segment durations, determining \MEL~and \ITB~as the optimal baseline at a 30-second segment duration. The addition of DG to this baseline yields a \impAcc\% increase in accuracy and \impFscore\% in F1-score, achieving \accDG\% accuracy, \precDG\% precision, \recDG\% recall, and \FscoreDG\% F1-score, surpassing all existing benchmarks. Extensive ablation studies assess the impact of multimodal fusion, deep architectural choices, and DG, highlighting their combined contribution to robust and generalizable depression detection.
\end{abstract}


\section{Introduction}
\label{sec:introduction}



    

\begin{figure}[!t]
    \centering
    \includegraphics[width=0.97\linewidth]{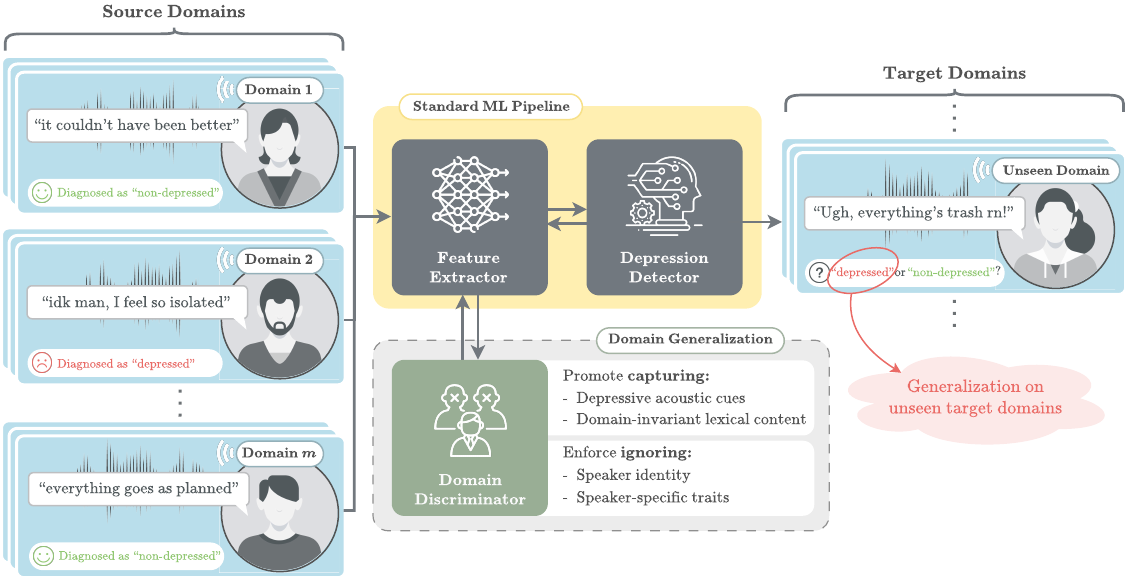}
    \caption{Overview of the proposed DG framework for multimodal depression detection. Each speaker is represented as a distinct \textit{domain} to model unique variability in audio-textual patterns. The \textit{domain discriminator} constrains the representations learned by the \textit{feature extractor} to suppress speaker-specific traits while retaining depression-related cues. This adversarial setup promotes domain-invariant features and enhances generalization to unseen domains.}
    \label{fig:pipeline}
\end{figure}

\IEEEPARstart{D}{epression}, one of the most prevalent mental health disorders worldwide~\cite{shorey2022global}, can progress to severe outcomes such as self-harm or suicide if left untreated~\cite{baldini2025suicidal}. This risk underscores the urgent need for reliable automated screening methods to complement existing preventive interventions~\cite{xiaMuLHiTANovelMulticlass2023}.

To enable automatic depression detection, recent advances in Artificial Intelligence (AI) have motivated systematic efforts to collect and analyze behavioral and physiological data, including speech, language, facial expressions, and biological signals, as distinctive indicators of depressive states~\cite{wu2023automatic, quDisentanglementProsodyRepresentations2025, yuanDiscoverySharedLatent2025}. Such data are then typically processed via feature extraction and modeling phases to uncover depression-related patterns.

Among these modalities, \textit{speech} and \textit{linguistic} features have proven to be particularly informative biomarkers of depression. Specifically, depressed patients often exhibit recurring acoustic patterns~\cite{koops2023speech}, such as reduced pitch variability, slower speaking rate, and longer pauses~\cite{di2025unraveling}, as well as linguistic traits like self-focused expressions and negative sentiments~\cite{gu2025linguistic}.

\IEEEpubidadjcol

Although standard audio-textual machine learning (ML) models have demonstrated improved performance in detecting depression~\cite{jabeen2023review}, a key barrier to their clinical deployment remains: the lack of subject-independent generalizability. Each individual exhibits unique speech characteristics and language use, which can cause models to capture personal behavioral markers rather than pathological indicators of depression. Consequently, despite strong performance on seen subjects, these models often struggle to generalize to new patients due to inter-speaker variability~\cite{quinonero2022dataset}.

To ensure reliability in real-world screening, diagnostic models must be resilient to such variations. Domain adaptation (DA)~\cite{farahani2021brief} and domain generalization (DG)~\cite{zhou2022domain} are two approaches addressing performance degradation under \textit{domain shift}~\cite{jahanifarDomainGeneralizationComputational2025}, which occurs when training data (\textit{source domains}) differ from that encountered during inference (\textit{target domains}). While DA requires access to target data during training, DG relies solely on source domains to simulate domain shifts, making it particularly suited to enhancing robustness when diagnosing unseen patients in depression detection~\cite{shenOutOfDistributionGeneralizationSurvey2021}.

In order to implement the DG paradigm, as shown in Fig.~\ref{fig:pipeline}, we treat each patient's unique audio-textual data as a separate \textit{domain}, inspired by~\cite{shankar2018generalizing}. Employing this perspective allows us to model the domain shift as inter-speaker variability, with source and target domains corresponding to patients in the training and test sets, respectively.

While a variety of DG methods have been explored in the literature, \textit{domain-invariant representation learning} techniques are extensively studied and proven effective~\cite{wangGeneralizingUnseenDomains2023}. In particular, these approaches encourage the learning of representations that ignore domain-specific confounding factors while preserving the discriminative patterns of the target task~\cite{rohlfsGeneralizationNeuralNetworks2025}.

Within this category, we adopt \textit{domain-adversarial training of neural networks (DANN)} proposed by Ganin \textit{et al.}~\cite{ganin2016domain} to guide the standard ML pipeline. Our framework incorporates a \textit{domain discriminator} equipped with a \textit{gradient reversal layer (GRL)}~\cite{ganin2016domain}, which penalizes the model when it correctly identifies the speaker. This adversarial mechanism drives the feature extractor to learn representations that are invariant to speaker-specific traits, yet remain pathologically informative. It thus facilitates a privacy-preserving, participant-independent depression detector that generalizes to unseen patients.

The proposed framework leverages a \textit{multi-source} DG paradigm~\cite{zhou2022domain}, jointly modeling acoustic and textual features via a \textit{model-based} fusion strategy~\cite{khoo2024machine, li2024enhancing}, all encapsulated within a unified architecture. Our key contributions include:

\begin{itemize}
    \item Proposing a novel multimodal audio-textual architecture with BiLSTM encoders, intra- and cross-modal attention modules, and segment-level decision-making.
    
    \item Conducting a comprehensive analysis of different feature extractor pairs for audio \textit{(\MEL, \HUB, \WAV)} and text \textit{(\BERT, \ITB, \ROB)} alongside varying segment durations.

    \item Evaluating the relative contribution of each modality and architectural component with extensive ablation studies.

    \item Achieving state-of-the-art results of \accDG\% accuracy and \FscoreDG\% F1-score after incorporating DG, outperforming prior benchmarks despite using 20\% less training data.
\end{itemize}

The rest of this paper is organized as follows:
Section \ref{sec:related} reviews prior studies for comparison, summarizes multimodal approaches, and highlights the relevance of DG.
Section~\ref{sec:formulation} formalizes the problem, providing an introductory context for Section~\ref{sec:methodology}, which details different modules of the proposed methodology.
Section~\ref{sec:setup} presents the experimental setup with a standardized protocol, followed by Section~\ref{sec:results}, which analyzes the results and presents a thorough ablation study assessing the proposed approach from diverse perspectives.
Finally, Sections \ref{sec:conclusion} and \ref{sec:future} summarize our conclusions and outline directions for future work.

\section{Related Work}
\label{sec:related}

This section summarizes prior depression detection studies as a reference for comparison, provides a brief but informative overview of multimodal depression detection, and discusses the relevance of domain generalization in this context.

{
\setlength{\tabcolsep}{4pt}
\begin{table}[!t]

\caption{Overview of depression classification studies on the Androids-Corpus. Some studies used \textit{reading ($R$)} or \textit{interview ($I$)} data exclusively, while others reported results for both in fusion $(R \land I)$ or separately $(R \lor I)$, adopting speech-only ($S$) or multimodal ($M$) approaches. Highlighted rows are not comparable to our study as we solely use the interview data.}
\label{tab:related-comparison}
\centering
\begin{tabular}{lccc}
    \toprule
        \textbf{Features (Model)} & \textbf{Data} & \textbf{Modality} \\
    \midrule
        OpenSMILE (SVM, LSTM) \cite{tao2023androids}  & $R \lor I$ & $S$ \\
         x-vector\,+\,emoHuBERT\,+\,TRILLsson (PTM) \cite{phukan2024avengers} & $R \lor I$ & $S$ \\
        MFCC\,+\,TEO\,+\,periodicity (HMM\,+\,MAP) \cite{ntalampiras2025interpretable} & $R \lor I$ & $S$ \\
    \midrule
        \rowcolor{gray!20} OpenSMILE (LSTM-CDMA) \cite{tao2024cross} & $R \land I$ & $S$ \\
        \rowcolor{gray!20} AlexNet (Mixture of Experts) \cite{ilias2025mixture} & $R \land I$
        & $S$ \\
        \rowcolor{gray!20} MFCC (Adaptive Knowledge Fusion) \cite{zhouAdaptiveKnowledgeFusion2025} & $R \land I$
        & $S$ \\
        \rowcolor{gray!20} STFT spectrograms\,+\,ResNet-18 (TFCA) \cite{rezaeeDepressionDetectionSpeech2026} & $R \land I$
        & $S$ \\
        \rowcolor{gray!20} OpenSMILE (CNN-LSTM) / TF-IDF (BERT) \cite{dalyDepressionDetectionRead2025} & $R \land I$
        & $M$ \\
    \midrule
        \rowcolor{gray!20} OpenSMILE correlations (SVM, LSTM) \cite{tao2023relationship} & $R$ & $S$ \\
        \rowcolor{gray!20} eGeMAPS, TRILLsson4 (XGBoost) \cite{polle2024revealing} & $R$ & $S$ \\
        \rowcolor{gray!20} OpenSMILE (CNN-BiLSTM\,+\,RL) \cite{liFAD3QNBrainInspiredDeep2025} & $R$ & $S$ \\
    \midrule
         PDEM embeddings (KELM) \cite{yu2025using} & $I$ & $S$ \\
         HuBERT, Wav2Vec2, Whisper (LR) \cite{de2024probing} & $I$ & $S$ \\
         DepAudioNet, ResNet, ECAPA-TDNN (MDFA) \cite{wangCrossLanguageDepressionDetection2025} & $I$ & $S$ \\
         DepAudioNet, ECAPA-TDNN (MTFS-Block) \cite{wangDepressionDetectionSpeech2026} & $I$ & $S$ \\
         Wav2Vec2 (CNN-GRU) / BERT (GRU) \cite{alsenani2024assessing} & $I$ & $M$ \\
         AlexNet / ItalianBERT (Cross-Attention) \cite{ilias2024cross} & $I$ & $M$ \\
         OpenSMILE, Wav2Vec2 / LIWC (FFN\,+\,MV) \cite{alsarraniPunctualContinuousAnalyzing2025} & $I$ & $M$ \\
         Wav2Vec2 / XLM-RoBERTa (D-CoPE\,+\,QF) \cite{zhangMitigatingInterviewerBias2025} & $I$ & $M$ \\
         
    \bottomrule
\end{tabular}
\end{table}
}

\subsection{Depression Detection on Androids-Corpus}\label{related:AC}

For comparative purposes in Section~\ref{subsec:results_AC}, here we present prior depression detection works on the Androids-Corpus~\cite{tao2023androids} dataset (see Section~\ref{sec:setup_dataset}). This dataset contains recordings of \textit{interview} ($I$) and \textit{reading} ($R$) tasks, prompting researchers to adopt a variety of strategies, using these data types either independently or in combination, as outlined in Table~\ref{tab:related-comparison}.

Some studies reported depression detection results on each data type separately ($R \lor I$). The initial baselines by Tao \textit{et al.}~\cite{tao2023androids} utilized OpenSMILE features, with \textit{BS1} employing an SVM and \textit{BS2} an LSTM with segment-level majority voting. Phukan \textit{et al.}~\cite{phukan2024avengers} proposed \textit{FuSeR}, a deep fusion model with x-vector, TRILLsson, and emoHuBERT embeddings, whereas Ntalampiras~\cite{ntalampiras2025interpretable} achieved competitive performance using an interpretable HMM-based approach with MFCCs, TEO autocorrelation envelopes, and periodicity features.

Conversely, some studies integrated both data types in their experiments ($R \land I$). Tao \textit{et al.}~\cite{tao2024cross} developed \textit{Cross-Data Multilevel Attention (CDMA)}, employing attention-enabled LSTMs over OpenSMILE features. Ilias \textit{et al.}~\cite{ilias2025mixture} stacked log-Mel spectrograms with their first- and second-order derivatives into a 3D representation, subsequently extracting features via AlexNet and processing them with a \textit{Mixture of Experts (MoE)} architecture. Daly and Olukoya~\cite{dalyDepressionDetectionRead2025} processed OpenSMILE features using a hybrid CNN-LSTM while encoding TF-IDF transcripts with BERT, fusing modality-specific outputs at the decision level. Zhou \textit{et al.}~\cite{zhouAdaptiveKnowledgeFusion2025} utilized an \textit{Adaptive Knowledge Fusion (AKF)} framework that jointly models temporal and channel-domain information from MFCC features. A similar study by Rezaee~\cite{rezaeeDepressionDetectionSpeech2026} combined a ResNet-18 backbone and a \textit{Temporal-Frequency-Channel Attention (TFCA)} mechanism to process STFT spectrogram segments.

Other investigations focused solely on a specific data type. Using the reading data ($R$), Tao \textit{et al.}~\cite{tao2023relationship} predicted depression with OpenSMILE-derived correlation representations, whereas Polle \textit{et al.}~\cite{polle2024revealing} fed eGeMAPS and TRILLsson4 features into XGBoost to identify confounding biases. With a brain-inspired reinforcement learning architecture, Li \textit{et al.}~\cite{liFAD3QNBrainInspiredDeep2025} processed OpenSMILE features via 1-D CNN and BiLSTM modules.

Using the interview data ($I$), Yu \textit{et al.}~\cite{yu2025using} fine-tuned a Wav2Vec2-based \textit{PDEM} model and used a \textit{Kernel Extreme Learning Machine (KELM)} for classification, while de Gennes \textit{et al.}~\cite{de2024probing} benchmarked HuBERT, Wav2Vec2, and Whisper embeddings with logistic regression. Alsenani \textit{et al.}~\cite{alsenani2024assessing} processed Wav2Vec2 audio and BERT text embeddings within a GRU network. Similar to~\cite{ilias2025mixture}, Ilias \textit{et al.}~\cite{ilias2024cross} extracted 3D AlexNet-based log-Mel spectrograms and ItalianBERT text features, applying cross-attention instead. Among recent works, Alsarrani \textit{et al.}~\cite{alsarraniPunctualContinuousAnalyzing2025} employed majority voting over audio (OpenSMILE or Wav2Vec2) and LIWC text features. Zhang and Poellabauer~\cite{zhangMitigatingInterviewerBias2025} proposed Dialogue-based CoPE (D-CoPE) on Wav2Vec2 and XLM-RoBERTa, leveraging an adversarial setup with GRL and dialogue-level transformers. Similarly, Wang \textit{et al.}~\cite{wangCrossLanguageDepressionDetection2025} proposed \textit{Multi-Domain Feature Alignment (MDFA)}, using GRL to learn domain-invariant representations from filter banks via different encoders. In their latest work~\cite{wangDepressionDetectionSpeech2026}, they used a \textit{Multiple Temporal–Frequency Scale Block (MTFS-Block)} and DWT to enhance DepAudioNet and ECAPA-TDNN baselines.

Most of these studies relied solely on acoustic features, overlooking potential gains from textual information. A further limitation is the lack of a dedicated \textit{validation} set in their experimental design. While~\cite{yu2025using} used nested cross-validation and~\cite{ntalampiras2025interpretable, rezaeeDepressionDetectionSpeech2026, zhouAdaptiveKnowledgeFusion2025} reserved a portion of the training data for validation, others trained for a fixed number of epochs, increasing the risk of overfitting and model selection bias.

\subsection{Multimodal Depression Detection}

Depression manifests through diverse behavioral signals, making multimodal approaches that integrate audio, text, and visual data well-suited for automated detection. By capturing complementary cues, such methods can typically outperform unimodal systems. Early research by Yang \textit{et al.}~\cite{yang2017multimodal} fused audio, video, and text descriptors using a deep CNN, while Haque \textit{et al.}~\cite{haque2018measuring} later applied a causal CNN to model acoustic, linguistic, and facial features. In a more recent study, Li \textit{et al.}~\cite{li2024enhancing} introduced \textit{IISFD}, achieving strong results by integrating visual feature extractors with acoustic and textual features through contrastive learning.

Notably, a major line of research has focused on bimodal audio-text fusion. Early work by Tuka \textit{et al.}~\cite{al2018detecting} modeled audio and text with separate LSTMs, merging the resulting representations in a feedforward network. Later, Makiuchi \textit{et al.}~\cite{rodrigues2019multimodal} estimated depression status by fusing BERT-based text embeddings with speech features extracted via a pretrained VGG-16 network, further processed using CNN, gated CNN, and LSTM layers. Toto \textit{et al.}~\cite{toto2021audibert} proposed \textit{AudiBERT}, a dual self-attentive BiLSTM framework integrating BERT with pretrained audio encoders such as VGGish, SincNet, or Wav2Vec. Shen \textit{et al.}~\cite{shen2022automatic} used a GRU-based network on audio Mel spectrograms and a BiLSTM on sentence embeddings, combined through modal attention to predict depression. More recently, Jia \textit{et al.}~\cite{jia2024bidirectional} introduced the \textit{bidirectional multimodal block-recurrent transformer (BMBRT)}, combining BERT and HuBERT for text and audio, respectively. Ding \textit{et al.}~\cite{ding2024intervoxnet} then presented \textit{IntervoxNet}, integrating a hybrid BERT-CNN text encoder with \textit{Audio Mel-Spectrogram Transformer (AMST)}. Employing BERT for text and Mel spectrograms for audio, Chen \textit{et al.}~\cite{chen2025text} achieved promising results with a BiLSTM, cross-attention, and transformer-based fusion network.

The demonstrated success of audio-textual approaches~\cite{ding2024intervoxnet, jia2024bidirectional} and the effectiveness of intra- and cross-modal attention modules~\cite{li2024enhancing, ilias2024cross} motivated us to adopt a similar strategy, yielding an end-to-end framework that serves as a comparative baseline prior to integrating DG.

\subsection{Domain Generalization}

In general, DG has been extensively studied in computer vision, medical imaging, and natural language processing, but its application to speech or multimodal tasks remains limited~\cite{wangGeneralizingUnseenDomains2023, zhou2022domain}. In the context of multimodal DG, Zhang \textit{et al.}~\cite{zhang2023video} introduced the \textit{DeVADG} framework, which addresses video-audio domain generalization from a causal perspective by disentangling confounding factors. Planamente \textit{et al.}~\cite{planamente2022domain} proposed a novel audio-visual loss function to balance the contributions of both modalities across domains by aligning their feature norms. Dong \textit{et al.}~\cite{dong2023simmmdg} developed \textit{SimMMDG}, a contrastive learning approach on the modality-shared features with distance constraints on modality-specific representations to encourage diversity. In another study, Dong \textit{et al.}~\cite{dong2024towards} presented \textit{MOOSA}, which tackles multimodal open-set domain generalization through self-supervised learning.

Despite these advances, multimodal speech-based DG for depression detection remains largely unexplored. Similar to our approach, Wang \textit{et al.}\cite{wangCrossLanguageDepressionDetection2025} and Liu \textit{et al.}\cite{liu2025domain} employed GRL to learn domain-invariant representations, but the former focused on unimodal DA, and the latter conducted unimodal DG using EEG rather than speech. Zhang and Poellabauer~\cite{zhangMitigatingInterviewerBias2025} is the closest work to ours, applying GRL for multimodal DG. However, they aim to mitigate bias arising from the inclusion of the interviewer's transcripts, while our adversarial setup targets depression-specific features robust to the inter-speaker variability. Unlike previous works, we investigate multimodal DG specifically for patient-independent depression detection, bridging a critical gap in the literature.
    
\section{Problem Formulation}
\label{sec:formulation}

Consider a dataset with an arbitrary number $N_i$ of variable-length audio recordings collected from each speaker $i$ during an interview. These recordings are then preprocessed into $K_i$ fixed-size audio segments, and the corresponding transcripts are obtained. Let $\mathcal{X}$ denote the joint audio-textual feature space and $\mathcal{Y}$ the binary label space for depression. We treat each participant as a distinct \textit{domain}, for whom a joint distribution $P_{XY}$ over $\mathcal{X} \times \mathcal{Y}$ is observed. Given $m$ \textit{source} domains defined as $\{\mathcal{D}_i^{(\mathcal{S})}\}_{i=1}^{m}$, each domain can be represented as:
\begin{equation}
    \label{eq:domain}
    \mathcal{D}_i^{(\mathcal{S})} = \left\{(\mathbf{x}_{i,k}^{(a)}, \mathbf{x}_{i,k}^{(t)}, d_i, y_i)\right\}_{k=1}^{K_i}
\end{equation}
\noindent where $\mathbf{x}_{i,k}^{(a)}$ and $\mathbf{x}_{i,k}^{(t)}$ represent the \textit{acoustic} and \textit{textual} features extracted by a \textit{modality-specific feature extractor} for segment $k$ of the $i$-th domain; $d_i \in \{1,\dots,m\}$ is the source domain label; and $y_i \in \{0,1\}$ indicates the depression status.

During training, a deep neural network referred to as the \textit{multimodal feature extractor ($f_\theta$)} maps each audio-textual pair to a latent representation $\mathbf{z}_{i,k} = f_\theta(\mathbf{x}_{i,k}^{(a)}, \mathbf{x}_{i,k}^{(t)}) \in \mathbb{R}^d$. This representation is expected to preserve depression-related cues while suppressing speaker-specific characteristics. To this end, the following modules are introduced:
\begin{itemize}
    \item A \textit{depression detector} ($h_\phi$), predicting depression logits $\hat{y}_{i,k} = h_\phi(\mathbf{z}_{i,k}) \in \mathbb{R}$. The detection error is measured via binary cross-entropy, such that computing $\mathrm{BCE}(\hat{y}_{i,k}, y_i)$ across all $K_i$ segments of $m$ source domains yields the overall depression loss $\mathcal{L}_{\mathrm{dep}}(\theta, \phi)$. Consequently, this loss is \textit{minimized} with respect to both $\theta$ and $\phi$, enabling more accurate depression detection.

    \item A \textit{domain discriminator} ($g_\psi$), predicting domain logits $\hat{d}_{i,k} = g_\psi(\mathbf{z}_{i,k}) \in \mathbb{R}^{m}$. The overall discrimination error $\mathcal{L}_{\mathrm{dom}}(\theta, \psi)$ is defined by computing the cross-entropy loss $\mathrm{CE}(\hat{d}_{i,k}, d_i)$ across all $K_i$ segments of $m$ source domains. While $g_\psi$ aims to \textit{minimize} $\mathcal{L}_{\mathrm{dom}}$ to better identify domains, $f_\theta$ \textit{maximizes} it adversarially, tricking $g_\psi$ so that it fails to capture domain-specific features.
\end{itemize}
This adversarial training strategy promotes domain-invariant representations, thereby enabling generalization to $n$ unseen target domains $\{\mathcal{D}_j^{(\mathcal{T})}\}_{j=1}^{n}$.
    \section{The Proposed Methodology}
\label{sec:methodology}

This section presents an overview of the entire pipeline, elaborating on the key components introduced in Section~\ref{sec:formulation} and their interplay, specifically the
\begin{inparaenum}[a)]
\item preprocessing module, 
\item audio-text feature extractor specifications,
\item multimodal feature extractor architecture, 
\item domain-adversarial training, and 
\item inference procedure.
\end{inparaenum}

\subsection{Preprocessing and Transcript Extraction}

\begin{figure}[!t]
    \centering
    \includegraphics[width=\linewidth]{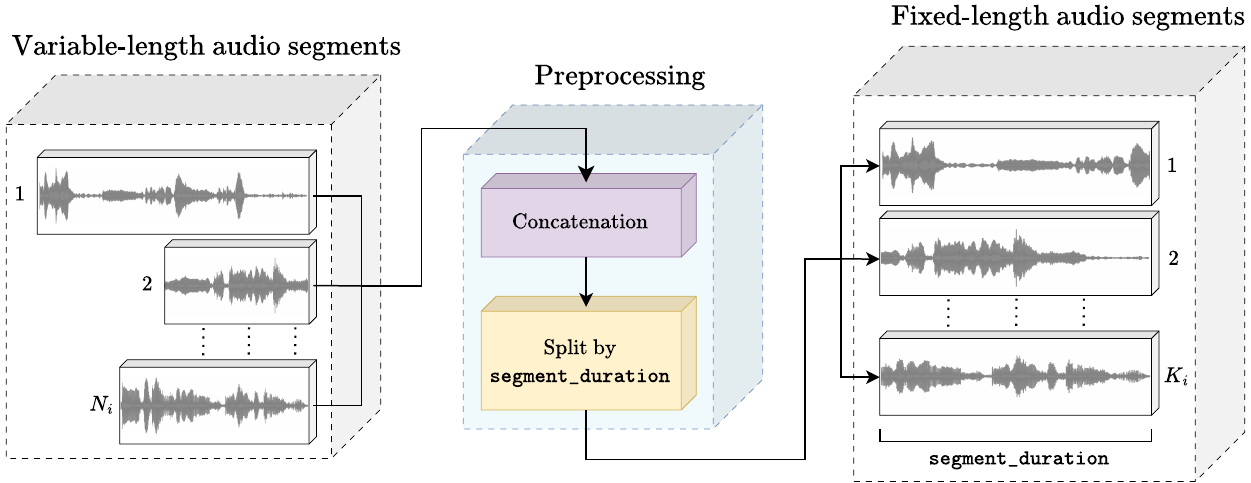}
    \caption{Preprocessing strategy: Converting $N_i$ variable-length audio segments of the $i$-th domain into $K_i$ fixed-length segments of a predefined duration.}
    \label{fig:preprocessing}
\end{figure}

As shown in Fig.~\ref{fig:preprocessing}, each domain comprises $N_i$ disjoint variable-length audio segments, which are first concatenated in chronological order to reconstruct a continuous waveform representing the participant's full speech during the interview. This waveform is then split into $K_i$ segments of a chosen duration, standardizing the input to a fixed number of samples \textit{(segment duration $\times$ sample rate)}. To address the variability in total interview length across domains, if the last split is shorter than 25\% of the target duration, it is discarded; otherwise, it is zero-padded to match the intended segment duration.

Each resulting fixed-length audio segment is transcribed using the \textit{``whisper-large-v3''} model, yielding a corresponding textual representation. This process ensures alignment between the acoustic signal and its transcript at the segment level, facilitating effective multimodal analysis.

\subsection{Modality-Specific Feature Extractors}\label{sec:method_modality_features}
To identify the optimal audio-textual feature extractor pair, different modality-specific feature extractors were evaluated: \textit{\{\MEL, \HUB, \WAV\}}~for audio, alongside \textit{\{\BERT, \ITB, \ROB\}}~for text. Aside from \MEL, all transformer models were implemented via Hugging Face, leveraging their last hidden state as frame-level embeddings for each segment. These strategies are detailed in the following.

\subsubsection{Audio Feature Extractor}
Regardless of the method, the extracted acoustic features are standardized into a tensor $\mathbf{x}_{i,k}^{(a)} \in \mathbb{R}^{L^{(a)} \times H^{(a)}}$ for each segment $k$ of the $i$-th participant, where $L^{(a)}$ denotes the number of frames per segment, and $H^{(a)}$ represents the frame-level feature embedding dimension. In particular, the following methods are evaluated:
\begin{itemize}
    \item \textit{\MEL:} Log-Mel spectrograms with FFT size \nFFT, hop length \hopLength, and \nMels~filter banks, with first- and second-order derivatives concatenated along the feature axis $(H^{(a)} = 128 \times 3 = 384)$.    
    \item \textit{\HUB:} \textit{``facebook/hubert-large-ls960-ft''} as the large variant of HuBERT~\cite{hsu2021hubert} $(H^{(a)} = 1024)$.
    \item \textit{\WAV:} \textit{``facebook/wav2vec2-base-960h''} as the base variant of Wav2Vec2~\cite{baevski2020wav2vec} $(H^{(a)} = 768)$.
\end{itemize}

\subsubsection{Text Feature Extractor}
The extracted textual features are represented as $\mathbf{x}_{i,k}^{(t)} \in \mathbb{R}^{L^{(t)}_i \times H^{(t)}}$, where $H^{(t)}$ denotes the token embedding dimension, but $L^{(t)}_i$ is the participant-specific maximum sequence length, to which shorter sequences are zero-padded, since the number of spoken tokens varies across segments of participant $i$. These embeddings are generated using one of the following pretrained models:
\begin{itemize}
    \item \textit{\BERT:} \textit{``google-bert/bert-base-multilingual-cased''} as the multilingual variant of BERT~\cite{devlin2019bert} $(H^{(t)} = 768)$.
    \item \textit{\ITB:} \textit{``dbmdz/bert-base-italian-xxl-cased''} as the Italian-specific variant of BERT~\cite{italianbert} $(H^{(t)} = 768)$.
    \item \textit{\ROB:} \textit{``FacebookAI/xlm-roberta-large''} as the multilingual variant of RoBERTa~\cite{conneau2019unsupervised} $(H^{(t)} = 1024)$.
\end{itemize}
\noindent In short, the selected audio-text feature extractor pair converts the preprocessed inputs into $(\mathbf{x}_{i,k}^{(a)}, \mathbf{x}_{i,k}^{(t)})$ for subsequent steps.%
\subsection{Multimodal Feature Extractor ($f_\theta$)}

\begin{figure*}[!t]
    \centering
    \includegraphics[width=\textwidth]{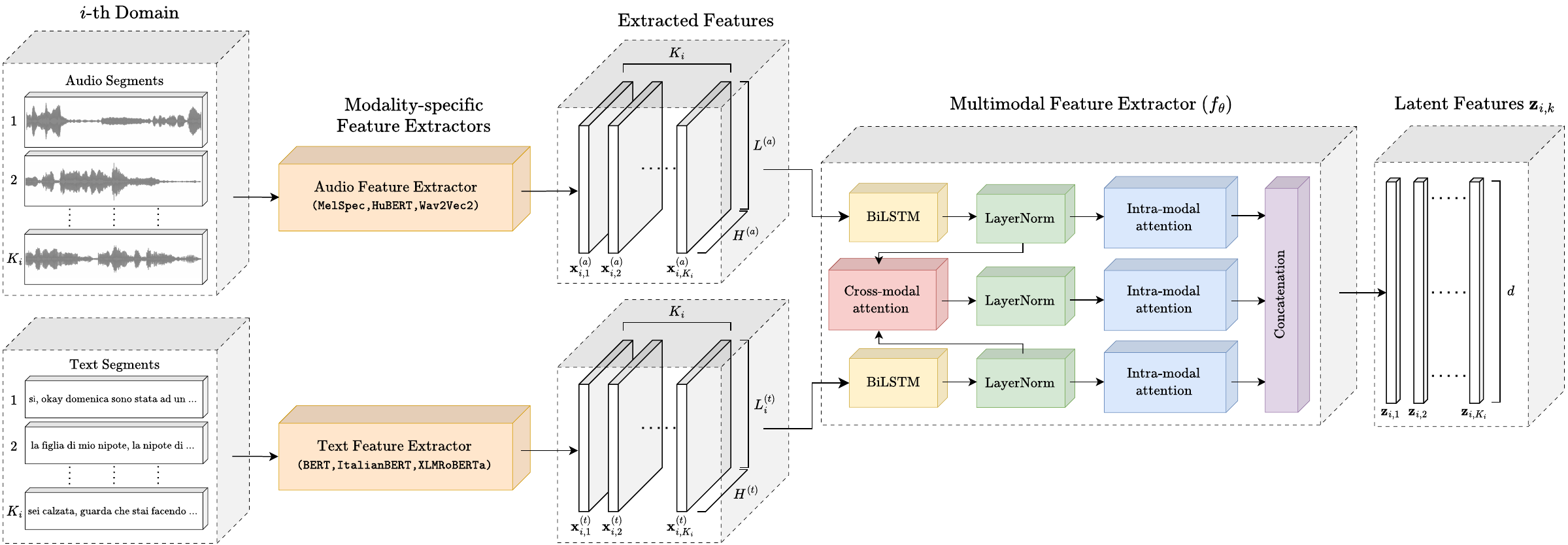}
    \caption{Feature extraction strategy: For each domain $i$, the $K_i$ fixed-length audio segments and corresponding transcripts are first fed into their \textit{modality-specific feature extractors} to obtain primary features. Our \textit{multimodal feature extractor} further processes these features, capturing temporal dependencies via BiLSTM and encoding interactions via attention modules. The resulting latent features are then used in an adversarial setup to learn domain-invariant representations.}
    \label{fig:feature_extraction}
\end{figure*}

This module maps each input pair $(\mathbf{x}_{i,k}^{(a)}, \mathbf{x}_{i,k}^{(t)})$ to a latent feature $\mathbf{z}_{i,k} \in \mathbb{R}^d$ through a structured pipeline of specialized sub-modules. As illustrated in Fig.~\ref{fig:feature_extraction}, an independent BiLSTM is first applied to each modality to capture both forward and backward temporal dependencies. This bidirectional approach ensures that the evolving context within both the frame-level acoustic sequences and the token-level textual features is fully captured. Following layer normalization to stabilize training, intra-modal attention highlights distinctive temporal segments within each modality, whereas cross-modal attention captures complementary audio-textual interactions. Together, the core components of $f_\theta$ yield the latent representations $\mathbf{z}_{i,k}$, with their architectural configurations detailed in the following.

\subsubsection{BiLSTM module}
Each includes a single hidden layer of size \lstmHiddenDim, concatenating forward and backward sequences and passing them through a normalization layer. Given $(\mathbf{x}_{i,k}^{(a)}, \mathbf{x}_{i,k}^{(t)})$, the output sequences are:
\begin{equation}\label{formula:seq_audio_text}
\begin{aligned}
\mathbf{seq}_{i,k}^{(a)} &= 
\mathrm{LayerNorm}\bigl(\mathrm{BiLSTM}^{(a)}(\mathbf{x}_{i,k}^{(a)})\bigr)
\in \mathbb{R}^{L^{(a)} \times \lstmHiddenDimTimesTwo} \\
\mathbf{seq}_{i,k}^{(t)} &= 
\mathrm{LayerNorm}\bigl(\mathrm{BiLSTM}^{(t)}(\mathbf{x}_{i,k}^{(t)})\bigr)
\in \mathbb{R}^{L^{(t)}_i \times \lstmHiddenDimTimesTwo}
\end{aligned}
\end{equation}

\subsubsection{Cross-modal attention (CMA)}
Let $\alpha = \mathrm{MHA}(Q, K, V)$ be the output of a multi-head attention layer with \nHeadsCrossAttn~heads and hidden size \crossAttnHiddenDim~that captures interactions among modalities, with the textual sequence as query attending to the aligned acoustic context ($Q = \mathbf{seq}_{i,k}^{(t)}$, and $K = V = \mathbf{seq}_{i,k}^{(a)}$). This output is then passed through a dropout layer ($p = \dropout$), added to the query, and normalized. Formally:
\begin{equation}\label{formula:seq_cross}
    \mathbf{seq}_{i,k}^{(c)} = \mathrm{LayerNorm}\bigl(Q + \mathrm{Dropout}(\alpha)\bigr) \in \mathbb{R}^{L^{(t)}_i \times \crossAttnHiddenDim}
\end{equation}

\subsubsection{Intra-modal attention (IMA)}
A linear attention network with a single hidden layer of size \attnHiddenDim~and \textit{Tanh} activation, followed by \textit{Softmax} and a dropout layer ($p = \dropout$) will first compute the attention weights for a given sequence $s \in \mathbb{R}^{L \times H}$. These weights are then applied to the original sequence $s$ via a dot product, yielding a compact representation $v \in \mathbb{R}^{H}$ that captures its most salient information. We apply this module to the output sequences from Eqs.~\eqref{formula:seq_audio_text} and~\eqref{formula:seq_cross} as: 
\begin{equation}
\label{formula:IMA}
\begin{aligned}
    \mathrm{IMA}(s) &= \mathrm{Dropout}\bigl(\mathrm{Softmax}(\mathrm{Linear}(s))\bigr) \cdot s\\
    \mathbf{v}_{i,k}^{(\alpha)} &= \mathrm{IMA}\bigl(\mathbf{seq}_{i,k}^{(\alpha)}\bigr) \in \mathbb{R}^{H},\quad \alpha \in \{a, t, c\}
\end{aligned}
\end{equation}


\subsubsection{Joint feature representation}
The final representation is constructed by \textit{concatenating} the attended features in Eq.~\eqref{formula:IMA}, with $d = \lstmHiddenDimTimesTwo + \lstmHiddenDimTimesTwo + \crossAttnHiddenDim = \finalDimension$ representing the total hidden dimensionality contributed by each component:
\begin{equation}\label{eq:latent_features}
    \mathbf{z}_{i,k} = f_\theta\bigl(\mathbf{x}_{i,k}^{(a)}, \mathbf{x}_{i,k}^{(t)}\bigr) = \left[\mathbf{v}_{i,k}^{(a)};  \mathbf{v}_{i,k}^{(t)}; \mathbf{v}_{i,k}^{(c)} \right] \in \mathbb{R}^d
\end{equation}

\subsection{Domain-Adversarial Training}

\begin{figure*}[!t]
    \centering
    \includegraphics[width=\textwidth]{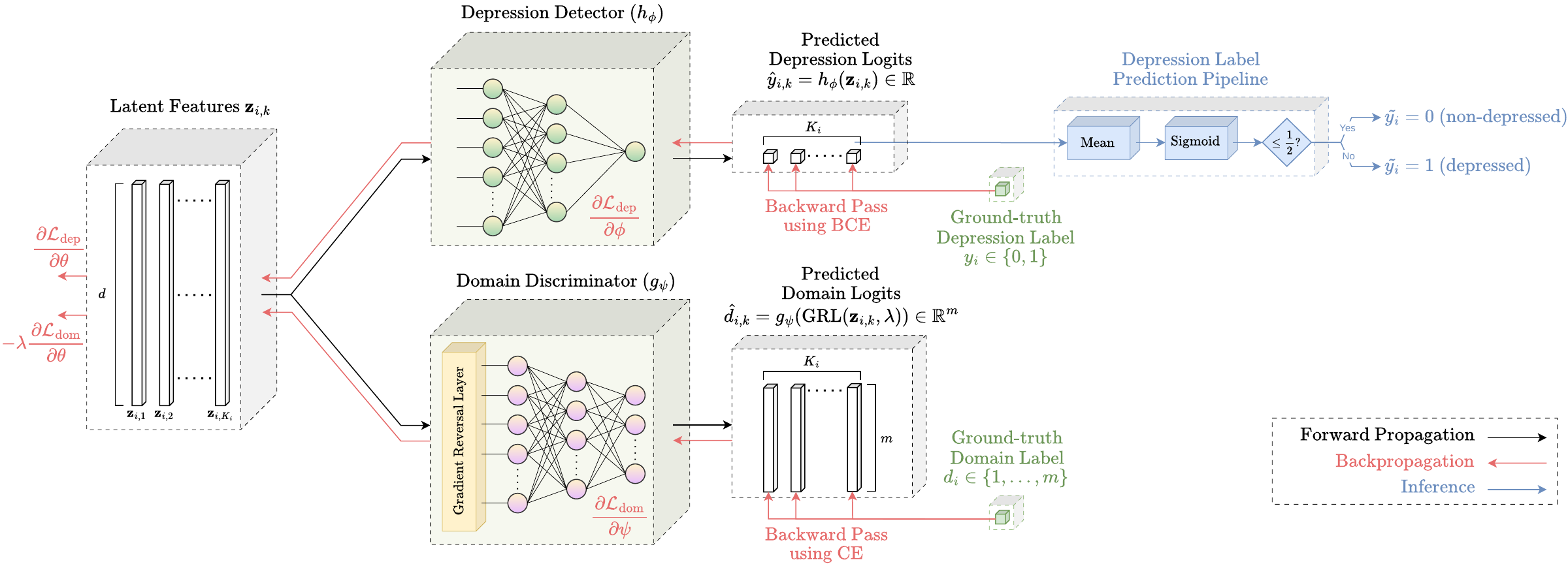}
    \caption{Adversarial framework between the depression detector ($h_\phi$) and the domain discriminator ($g_\psi$). A gradient reversal layer (GRL) inverts gradients from $g_\psi$ during backpropagation, enabling joint optimization of $\theta$, $\phi$, and $\psi$ within a single minimization objective.}
    \label{fig:adversarial}
\end{figure*}

After extracting the latent features $\mathbf{z}_{i,k}$ using the multimodal feature extractor ($f_\theta$) as defined in Eq.~\eqref{eq:latent_features}, the objective is to ensure that they preserve depression-related information while remaining invariant to inter-speaker variability. To this end, an adversarial framework between a \textit{depression detector} ($h_\phi$) and a \textit{domain discriminator} ($g_\psi$) is introduced, which is described in detail in the following.

\subsubsection{Depression detector ($h_\phi$)}
The depression detector is implemented as a fully connected network with a single hidden layer of \fcHiddenDim~neurons, followed by a \textit{ReLU} activation function and a single output neuron that estimates a depression logit for each segment $k$, defined as $\hat{y}_{i,k} = h_\phi(\mathbf{z}_{i,k}) \in \mathbb{R}$. The discrepancy between this predicted logit and the ground-truth label for domain $i$ is measured using the binary cross-entropy (BCE) loss during training, computed as:
\begin{equation}
    \small{\mathrm{BCE}}(\hat{y}_{i,k}, y_i) =
        -\bigl[ y_i \log \sigma(\hat{y}_{i,k}) 
        + (1 - y_i) \log (1 - \sigma(\hat{y}_{i,k})) \bigr]
\end{equation}
\noindent where $\sigma(z) = \frac{1}{1 + e^{-z}}$ denotes the sigmoid activation function.
Let $\mathcal{L}_{\mathrm{dep}}$ be the overall depression detection loss, defined as the average loss across all segments $k$ and source domains $m$:
\begin{equation}
    \mathcal{L}_{\mathrm{dep}}(\theta, \phi) 
    = \frac{1}{m} \sum_{i=1}^{m} 
        \left( \frac{1}{K_i} \sum_{k=1}^{K_i} \mathrm{BCE}(\hat{y}_{i,k}, y_i) \right)
\end{equation}
\noindent
In this setting, the optimization objective is to learn the model parameters $\theta$ and $\phi$ that \textit{minimize} $\mathcal{L}_{\mathrm{dep}}$, enabling accurate prediction of depression for individuals:
\begin{equation}\label{eq:depression_minimization}
    (\hat{\theta}, \hat{\phi}) \gets \operatorname*{argmin}_{\theta, \phi} \mathcal{L}_{\mathrm{dep}}(\theta, \phi)
\end{equation}

\subsubsection{Domain discriminator ($g_\psi$)}
This module is designed as another fully connected network with a single hidden layer of size \fcHiddenDim, followed by a \textit{ReLU} activation function and an output layer with $m$ neurons. Taking as input the same latent features $\mathbf{z}_{i,k}$ for each domain $i$, it produces $\hat{d}_{i,k} = g_\psi(\mathbf{z}_{i,k}) \in \mathbb{R}^{m}$ as an output vector over the $m$ possible source domains, such that $\hat{d}_{i,k}[j]$ denotes the logit corresponding to the $j$-th domain. During training, the prediction error of $g_\psi$ is quantified using the cross-entropy (CE) loss as:
\begin{equation}
    \mathrm{CE}(\hat{d}_{i,k}, d_i) = -\log \biggl( \frac{\exp(\hat{d}_{i,k}[d_i])}{\sum_{j=1}^{m} \exp(\hat{d}_{i,k}[j])} \biggl)
\end{equation}
where $\exp(z) = e^z$. The overall domain discrimination loss $\mathcal{L}_{\mathrm{dom}}$ is therefore obtained by averaging over all segments $k$ and source domains $m$:
\begin{equation}
    \mathcal{L}_{\mathrm{dom}}(\theta, \psi) 
    = \frac{1}{m} \sum_{i=1}^{m} 
        \left( \frac{1}{K_i} \sum_{k=1}^{K_i} \mathrm{CE}(\hat{d}_{i,k}, d_i) \right)
\end{equation}
\noindent
Recall that to enable DG, $g_\psi$ is expected to perform poorly at domain discrimination, indicating that $f_\theta$ has successfully removed domain-specific information from $\mathbf{z}_{i,k}$. In particular, $g_\psi$ \textit{minimizes} $\mathcal{L}_{\mathrm{dom}}$ to better distinguish source domains, whereas $f_\theta$ seeks to \textit{maximize} the same loss, preventing $g_\psi$ from achieving its goal. This adversarial update encourages domain-invariant features to emerge, yielding the following minimax optimization objective:
\begin{equation}\label{eq:minimax}
    (\hat{\theta}, \hat{\psi}) \gets \operatorname*{argmax}_{\theta} \operatorname*{argmin}_{\psi} \mathcal{L}_{\mathrm{dom}}(\theta, \psi)
\end{equation}
\noindent
To solve this non-trivial optimization via backpropagation, a GRL~\cite{ganin2016domain} is integrated into $g_\psi$, as shown in Fig.~\ref{fig:adversarial}. The domain logits are obtained as $\hat{d}_{i,k} = g_\psi(\mathrm{GRL}(\mathbf{z}_{i,k}, \lambda)) \in \mathbb{R}^{m}$, where $\lambda>0$ is a constant controlling the strength of domain confusion. During the forward pass, GRL acts as the identity function, while in backpropagation it multiplies gradients by $-\lambda$, effectively reversing their direction to $-\lambda \frac{\partial \mathcal{L}_{\mathrm{dom}}}{\partial \theta}$. This transformation enables reformulating Eqs.~\eqref{eq:depression_minimization} and~\eqref{eq:minimax} into a joint minimization objective:
\begin{equation}\label{formula:joint_objective}
    \min_{\theta, \phi, \psi}\, \mathcal{L}_{\mathrm{dep}}(\theta, \phi) + \mathcal{L}_{\mathrm{dom}}(\theta, \psi)
\end{equation}
\noindent
The full training pseudocode is presented in Algorithm~\ref{alg:DG_training}, detailing the joint optimization process that ultimately yields the parameters $\theta$ and $\phi$ used for inference.

\subsection{Inference}

To predict the final depression label for an individual \textit{target} domain, the decision relies solely on the domain-invariant representations learned by $f_\theta$ and $h_\phi$. Consequently, $g_\psi$ can be disabled at inference, as its role is limited to guiding feature learning during adversarial training. Given a target domain $i$, all its $K_i$ depression logits $\hat{y}_{i,k}$ obtained by the depression detector are averaged to obtain a single score:
\begin{equation}
    \bar{y_i} = \frac{1}{K_i} \displaystyle\sum_{k=1}^{K_i} \hat{y}_{i,k} = \frac{1}{K_i} \displaystyle\sum_{k=1}^{K_i} h_\phi\bigl(\overbrace{f_\theta(\mathbf{x}_{i,k}^{(a)}, \mathbf{x}_{i,k}^{(t)})}^{\mathbf{z}_{i,k}}\bigr) \in \mathbb{R}
\end{equation}
\noindent and the final predicted depression label is then obtained as:
\begin{equation}
    \tilde{y_i} =
    \begin{cases}
        0~\text{(non-depressed)} & \text{if } \sigma\left(\bar{y_i}\right) \leq \frac{1}{2} \\
        1~\text{(depressed)} & \text{otherwise}
    \end{cases}
\end{equation}

    \section{Experimental Setup}
\label{sec:setup}

In this section, we describe the experimental framework used to evaluate the proposed multimodal depression detection approach. Specifically, we detail:
\begin{inparaenum}[a)]
\item the dataset, including its structure, distribution, and relevance;
\item the standardized cross-validation protocol ensuring reproducibility; and
\item the training setup, with selected hyperparameter configurations used in the experiments.
\end{inparaenum}

\subsection{Dataset}\label{sec:setup_dataset}

\textit{Androids-Corpus}~\cite{tao2023androids} is a relatively recent and public Italian dataset, especially noteworthy given the scarcity of depression detection resources in the Italian language. It comprises 228 audio recordings sampled at 16~kHz contributed by 118 native Italian speakers, divided into two groups: 112 recordings of \textit{reading} speech reciting the same text with a total duration of roughly 1 hour and 34 minutes, and 116 recordings of \textit{interview} speech answering a fixed set of open-ended questions spanning around a total of 7 hours and 24 minutes.

Motivated by its spontaneous nature, our study adopts only the interview data, which provides a more realistic setting for speech-based depression detection. Among the corresponding 116 participants, 64 were clinically diagnosed with \textit{depression}, while 52 served as \textit{non-depressed} or \textit{control} participants, with both groups demographically balanced in age and education.

\begin{figure}[!t]
    \centering
    \includegraphics[width=\linewidth]{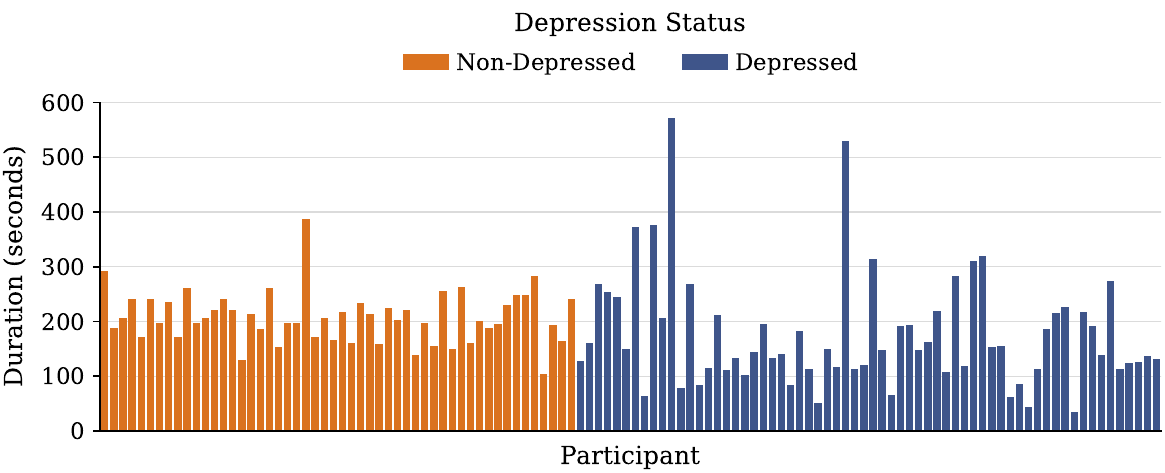}
    \caption{Waveform length distribution of participants in the Androids-Corpus after concatenating their respective \textit{interview} recordings.}
    \label{fig:durations}
\end{figure}

The dataset also provides a segmentation of the audio files to ease the process of eliminating the interviewer's speech. As shown in Fig.~\ref{fig:durations}, the duration of the concatenated recordings for each individual ranges approximately from 40 seconds to just under 600 seconds.

\begin{table}[!t]
\scriptsize
\caption{Overview of the hyperparameter grid explored in the experimental setup. The configuration adopted for the reported experiments is highlighted in bold.}
\label{tab:setup_hyperparams}
\centering
\begin{tabular}{lc}
\toprule
\textbf{Hyperparameter} & \textbf{Candidate Values} \\
\midrule
    \MEL~FFT window size & \{2048, \textbf{\nFFT}\} \\
    \MEL~hop length & \{256, \textbf{\hopLength}, 1024\} \\
    \MEL~number of Mel filter banks & \{\textbf{\nMels}, 192\} \\
    Audio BiLSTM hidden dimension & \{128, \textbf{\lstmHiddenDim}, 512\} \\
    Text BiLSTM hidden dimension & \{128, \textbf{\lstmHiddenDim}, 512\} \\
    Intra-modal attention hidden size & \{\textbf{\attnHiddenDim}, 256\} \\
    Cross-modal attention hidden size & \{128, \textbf{\crossAttnHiddenDim}, 512\} \\
    Cross-modal attention number of heads & \{\textbf{\nHeadsCrossAttn}\} \\
    Fully-connected layer hidden size & \{64, \textbf{\fcHiddenDim}, 192\} \\
    Dropout rate & \{\textbf{\dropout}, 0.2\} \\
    GRL scaling coefficient ($\lambda$) & \{0.3, \textbf{\lambdaGRL}, 0.7, 1\} \\
    Learning rate ($\eta$) & \{\textbf{1}, 2, 5\}$\mathbf{{\times}10^{-5}}$ \\
    AdamW optimizer weight decay & \{\textbf{1}, 2, 5\}$\mathbf{{\times}10^{-5}}$ \\
    ReduceLROnPlateau scheduler decay factor ($\gamma$) & \{0.1, 0.2, \textbf{\schedulerFactor}\} \\
    ReduceLROnPlateau scheduler patience epochs ($P_{\text{schd}}$) & \{\textbf{\schedulerPatience}, 2, 3, 5\} \\
    Early stopping epsilon ($\varepsilon$) & \{\textbf{\stopperEpsilon}, 0.2\} \\
    Early stopping patience epochs ($P_{\text{stop}}$) & \{2, \textbf{\stopperPatience}, 5\} \\
    Maximum training epochs ($N$) & \{\textbf{\numEpochs}\} \\
\bottomrule
\end{tabular}
\end{table}

\begin{algorithm}[!t]
\caption{MultimodalDG Training}\label{alg:DG_training}
\begin{algorithmic}[1]
\vspace{4pt}

\REQUIRE Learning rate $\eta$, GRL coefficient $\lambda$, max epochs $N$

\REQUIRE $m$ source domains $\{\mathcal{D}_i^{(\mathcal{S})}\}_{i=1}^{m}$, each comprising $K_i$ fixed-size segments $\{(\mathbf{x}_{i,k}^{(a)}, \mathbf{x}_{i,k}^{(t)}, d_i, y_i)\}_{k=1}^{K_i}$
\vspace{6pt}

\STATE Split $\{\mathcal{D}_i^{(\mathcal{S})}\}_{i=1}^{m}$ into training set $S$ and validation set $V$
\STATE \textbf{Initialize} randomly the model parameters $\theta$, $\phi$, $\psi$

\FOR{epoch $= 1$ \textbf{to} $N$}
    \STATE $\mathcal{L}_{\text{train}} \gets 0$, $\mathcal{L}_{\text{val}} \gets 0$
    
    \STATE
    \STATE \algcomment{\# Training Phase}
    \FOR{each domain $i$ \textbf{in} $S$}
        
        \FOR{segment $k = 1$ \textbf{to} $K_i$}
        
            \STATE Extract the latent features $\mathbf{z}_{i,k} \gets f_\theta(\mathbf{x}_{i,k}^{(a)}, \mathbf{x}_{i,k}^{(t)})$
            
            \STATE Predict depression logits $\hat{y}_{i,k} \gets h_\phi(\mathbf{z}_{i,k})$

            \STATE Predict domain logits $\hat{d}_{i,k} \gets g_\psi(\mathrm{GRL}(\mathbf{z}_{i,k}, \lambda))$

            \STATE $\ell_{i,k} \gets \mathrm{BCE}(\hat{y}_{i,k}, y_{i}) + \mathrm{CE}(\hat{d}_{i,k}, d_{i})$
        \ENDFOR

        \STATE Average the loss of segments $\mathcal{L}_{i} \gets \frac{1}{K_i} \sum_{k=1}^{K_i} \ell_{i,k}$
        \STATE Update $\theta, \phi, \psi$ to minimize $\mathcal{L}_{i}$ using $\mathrm{AdamW}(\eta)$
        
        \STATE $\mathcal{L}_{\text{train}} \gets \mathcal{L}_{\text{train}} + \mathcal{L}_{i}$
        
    \ENDFOR
    
    \STATE
    \STATE \algcomment{\# Validation Phase}
    \FOR{each domain $i$ \textbf{in} $V$}
        \FOR{segment $k = 1$ \textbf{to} $K_i$}
        
            \STATE Extract the latent features $\mathbf{z}_{i,k} \gets f_\theta(\mathbf{x}_{i,k}^{(a)}, \mathbf{x}_{i,k}^{(t)})$
            
            \STATE Predict depression logits $\hat{y}_{i,k} \gets h_\phi(\mathbf{z}_{i,k})$

            \STATE $\ell_{i,k} \gets \mathrm{BCE}(\hat{y}_{i,k}, y_{i})$
        \ENDFOR
        \STATE $\mathcal{L}_{\text{val}} \gets \mathcal{L}_{\text{val}} + \frac{1}{K_i} \sum_{k=1}^{K_i} \ell_{i,k}$
    \ENDFOR

    \STATE
    \IF{$\mathrm{EarlyStopping}(\mathcal{L}_{\text{train}}, \mathcal{L}_{\text{val}})$ is triggered}
        \STATE \textbf{break}
    \ENDIF
    
    \STATE Update $\eta$ using $\mathrm{ReduceLROnPlateau}(\mathcal{L}_{\text{val}})$
\ENDFOR
\STATE \textbf{return} $\theta, \phi$

\end{algorithmic}
\end{algorithm}

\subsection{$k$-fold Cross-Validation}

Replicating the data splits introduced by the authors of the dataset, the experiments in this study adopted a $k$-fold ($k=5$) cross-validation protocol. These splits keep all the participant's recordings in the same fold to ensure no overlap between train and test sets, which prevents memorizing speaker-specific traits. At each iteration $k = 1, \ldots, 5$, the $k$-th fold was treated as the test set, while the remaining data were further divided using a stratified split to maintain the distribution of labels across splits, with 80\% used for training and 20\% reserved for validation. As the dataset was almost balanced in terms of depressed vs non-depressed classes, standard evaluation metrics including accuracy, precision, recall, and F1-score were used for performance evaluation.

To ensure robustness and account for potential variability, the experimental results reported in Section~\ref{subsec:results_baseline} are presented as the mean and standard deviation over \textit{three} independent runs of each experiment. In contrast, the results in Sections~\ref{subsec:results_DG} and~\ref{subsec:results_ablation} are reported as the mean and standard deviation across folds from a single run, in line with prior studies using the Androids-Corpus to ensure fair comparison and reliable ablation analysis.

\subsection{Training Specifications}

Model training was performed using the \textit{AdamW} optimizer, with learning rate ($\eta$) and weight decay each set to \weightDecay. Using PyTorch's $\mathrm{ReduceLROnPlateau}$ scheduler with decay factor $\gamma=\schedulerFactor$ and patience $P_{\text{schd}}=\schedulerPatience$ epochs, $\eta$ was adjusted dynamically in response to plateaus in the validation loss.

As outlined in Algorithm~\ref{alg:DG_training}, training and validation losses were monitored using a custom $\mathrm{EarlyStopping}(\mathcal{L}_{\text{train}}, \mathcal{L}_{\text{val}})$ mechanism, which terminated training if:
\begin{inparaenum}[(i)]
    \item changes in $\mathcal{L}_{\text{train}}$ remained below $\varepsilon=\stopperEpsilon$ for $P_{\text{stop}}=\stopperPatience$ consecutive epochs, indicating convergence, or
    \item either $\mathcal{L}_{\text{train}}$ or $\mathcal{L}_{\text{val}}$ maintained an increasing trend over the same number of epochs.
\end{inparaenum}
Although the maximum number of epochs was set a priori to $N=\numEpochs$, training typically halted after no more than 35 epochs in most experiments, effectively preventing overfitting due to the rising validation loss.

Different hyperparameters in this study were chosen via empirical tuning over the search space summarized in Table~\ref{tab:setup_hyperparams}, with the final optimal values highlighted in bold for reference.

The proposed approach was developed using Python 3.12.8, built on the \textit{PyTorch} framework, and executed on a server with two \textit{NVIDIA TITAN V} GPUs (each with 12 GB of VRAM, running CUDA version 12.8).



\begin{table*}[!t]
\caption{Performance comparison of audio-text feature extractor pairings across different segment durations. Each cell reports Accuracy / Precision / Recall / F1-Score, presented in the same order as percentages (\%).}
\label{tab:results_comp_merged}
\centering
\begin{tabular}{cccccc}
\toprule
    \multirow{2}{*}{\parbox{1.8cm}{\centering \textbf{Audio Feature\\Extractor}}} &
    \multirow{2}{*}{\parbox{1.8cm}{\centering \textbf{Text Feature\\Extractor}}} &
    \multicolumn{4}{c}{\textbf{Segment Duration}} \\
    \cmidrule(lr){3-6}
     & & \textbf{20s} & \textbf{30s} & \textbf{45s} & \textbf{60s} \\
\midrule
    \multirow{3}{*}{\MEL}
    & \BERT & 77.2 / 76.2 / 88.1 / 80.5
            & 85.2 / 85.9 / 89.3 / 86.8
            & 84.3 / 84.9 / 87.3 / 85.4
            & 85.8 / 87.0 / \textbf{88.9} / 87.1 \\
    & \ITB  & 86.4 / \textbf{86.8} / 89.5 / 87.3
            & \textbf{90.4} / \textbf{90.8} / 92.0 / \textbf{90.8}
            & \textbf{88.1} / \textbf{87.1} / \textbf{90.9} / \textbf{88.7}
            & \textbf{87.5} / \textbf{90.4} / 87.8 / \textbf{88.0} \\
    & \ROB  & 68.5 / 66.1 / 92.7 / 76.1
            & 67.9 / 69.1 / 85.4 / 74.2
            & 70.8 / 69.8 / 85.6 / 75.0
            & 75.5 / 75.0 / 86.2 / 78.9 \\
\midrule
    \multirow{3}{*}{\HUB}
    & \BERT & 76.2 / 74.1 / 86.4 / 78.8
            & 77.9 / 76.9 / 88.0 / 80.9
            & 77.3 / 77.5 / 87.4 / 80.4
            & 79.9 / 81.1 / 85.4 / 82.0 \\
    & \ITB  & 83.1 / 80.4 / 91.7 / 85.0
            & 82.5 / 80.6 / 91.1 / 84.7
            & 85.4 / 85.7 / 89.5 / 86.6
            & 80.2 / 80.8 / 87.1 / 82.5 \\
    & \ROB  & 70.4 / 71.4 / 81.6 / 72.8
            & 71.3 / 72.4 / 87.7 / 76.0
            & 70.8 / 71.7 / 83.2 / 74.7
            & 76.4 / 75.5 / 86.5 / 78.6 \\
\midrule
    \multirow{3}{*}{\WAV}
    & \BERT & 78.8 / 77.6 / 86.6 / 80.9
            & 81.0 / 79.6 / 90.4 / 83.8
            & 79.7 / 84.5 / 82.8 / 80.4
            & 80.8 / 83.6 / 83.0 / 82.1 \\
    & \ITB  & \textbf{86.8} / 85.4 / \textbf{93.5} / \textbf{88.3}
            & 86.2 / 85.9 / \textbf{92.2} / 88.1
            & 83.1 / 85.5 / 87.7 / 85.1
            & 84.0 / 85.1 / 87.3 / 85.5 \\
    & \ROB  & 73.3 / 73.3 / 79.6 / 75.0
            & 73.9 / 75.0 / 82.7 / 76.6
            & 71.0 / 71.5 / 81.8 / 74.3
            & 81.6 / 81.1 / 86.1 / 82.7 \\
\bottomrule
\end{tabular}
\end{table*}
\section{Results}
\label{sec:results}

This section presents the experimental results of our study, specifically detailing:
\begin{inparaenum}[a)]
\item an analysis of feature extractors across segment durations,
\item the impact of domain generalization,
\item a comparison with state-of-the-art methods, and
\item ablation studies on modalities and architectural variants.
\end{inparaenum}

\subsection{Feature Extractors vs. Segment Durations}
\label{subsec:results_baseline}

In order to find a competitive baseline, different selections of modality-specific feature extractor pairs and segment durations for fixed-length audio segments were evaluated in a structured set of experiments. In particular, every possible pairing of three audio extractors (\MEL, \HUB, and \WAV) with three text extractors (\BERT, \ITB, and \ROB) across segment durations of 20, 30, 45, and 60 seconds were examined for this purpose.

As summarized in Table~\ref{tab:results_comp_merged}, the pairing of \MEL~feature extractor for audio and \ITB~for the text modality at a 30-second segment duration achieved the highest performance. With an average accuracy and F1-score of 90.4\% and 90.8\%, respectively, this combination was selected as the baseline for downstream analysis.

\begin{figure}[!t]
    \centering
    \includegraphics[width=0.94\linewidth]{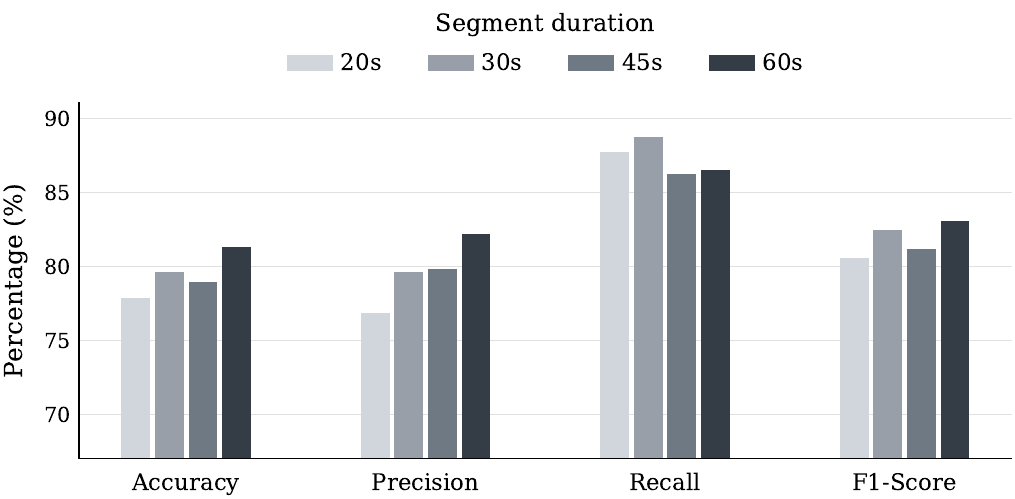}
    \caption{Average performance across all experiments for each segment duration, computed across all the audio-text pairings for each duration in Table~\ref{tab:results_comp_merged}.}
    \label{fig:perf_segments}
\end{figure}

To assess the impact of segment duration, Fig.~\ref{fig:perf_segments} presents average performance across all feature extractor combinations for each duration, derived from Table~\ref{tab:results_comp_merged}. Overall, adopting longer speech segments improves accuracy, though at the cost of higher GPU memory usage for processing larger sequences. Given this trade-off, a 30-second segment duration provides a good balance between performance and computational cost.

\begin{figure}[!t]
    \centering
    \includegraphics[width=0.94\linewidth]{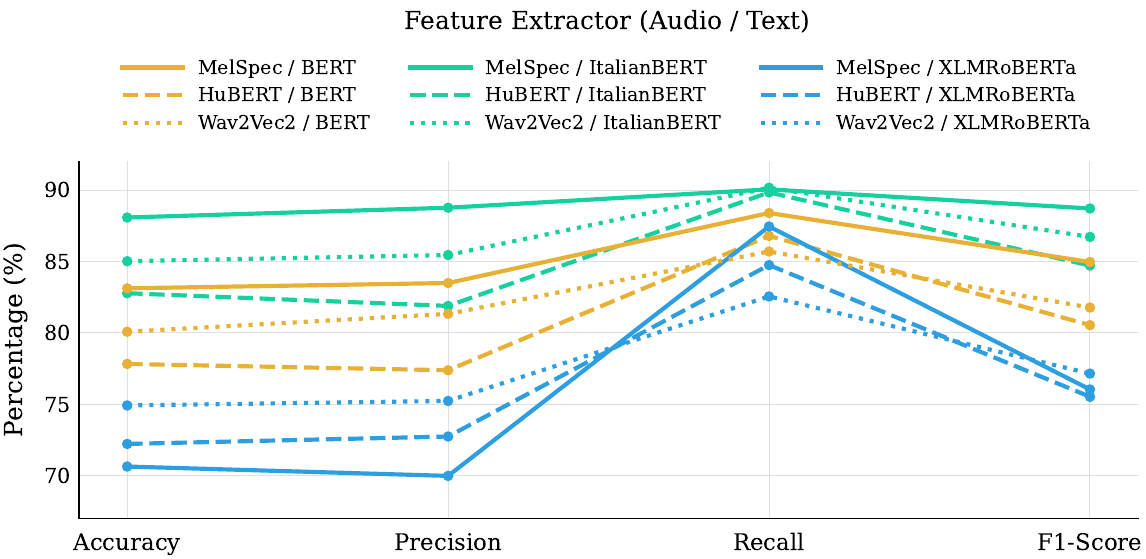}
    \caption{Average performance of modality-specific feature extractor pairings, computed across all segment durations for each pairing from Table~\ref{tab:results_comp_merged}.}
    \label{fig:perf_models}
\end{figure}

Alternatively, Fig.~\ref{fig:perf_models} illustrates the average results of each audio-textual feature extractor. The combination of \MEL\ (audio) and \ITB\ (text) consistently yields the highest performance, likely due to \MEL's compact representations and the Italian-specific design of \ITB.


\subsection{Domain Generalization Effect}
\label{subsec:results_DG}

The previously identified baseline model was extended by integrating the domain discriminator $g_\psi$ to assess the proposed DG framework. This inclusion led to an improvement of \impAcc\% in accuracy and \impFscore\% in F1-score, yielding a final performance of \accDG\% accuracy, \precDG\% precision, \recDG\% recall, and \FscoreDG\% F1-score, highlighting the contribution of adversarial training in improving generalization to unseen target domains.

Analyzing these performance gains, McNemar's test with continuity correction was applied to individual predictions. Given roughly 23 testing samples per fold in the standardized experimental protocol, achieving $p<0.05$ requires substantial disagreement between models. Hence, the null hypothesis was not rejected, reflecting that insufficient sample size constrains statistical significance, even with strong observed results.

\begin{figure}[!t]
    \centering
    \includegraphics[width=\linewidth]{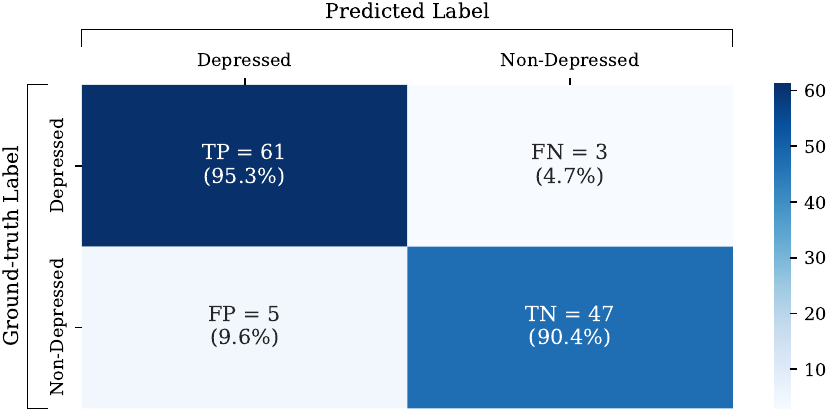}
    \caption{Cumulative confusion matrix of the proposed model, compiled by summing each fold's confusion matrix across the 5-fold cross-validation setup.}
    \label{fig:confusion_matrix}
\end{figure}

To provide an alternative perspective on model performance, we report the cumulative confusion matrix in Fig.~\ref{fig:confusion_matrix}, obtained by summing the per-fold confusion matrices across the 5-fold cross-validation. Unlike the averaged metrics reported earlier, this summation conceptually reflects the model's behavior over the entire dataset. The proposed model correctly identifies 95.3\% and 90.4\% of depressed and non-depressed individuals, respectively. In addition, it achieves a low false negative rate ($\text{FNR}=4.7$\%), which is crucial in clinical screening to avoid missing individuals at risk while maintaining strong specificity for the non-depressed group.

\subsection{Comparison with Androids-Corpus Benchmarks}\label{subsec:results_AC}

Table~\ref{tab:results_Dep} benchmarks the proposed approach against prior studies on the Androids-Corpus dataset (see Section \ref{related:AC} for reference). The proposed model outperforms all these efforts across every evaluation metric, attaining a peak F1-score of \FscoreDG\%, which represents a clear improvement over the 93.1\% of the closest competitor~\cite{alsarraniPunctualContinuousAnalyzing2025}. This comparison validates the strength and reliability of the proposed multimodal framework, and the effectiveness of the domain generalization strategy in capturing depression-specific feature embeddings.

\begin{table*}[!t]
\caption{Performance comparison with prior studies on the Androids-Corpus for depression detection, reported as percentages (\%).}
\label{tab:results_Dep}
\centering
\begin{tabular}{lcccccc}
    \toprule
    \textbf{Study} & \textbf{Model} & \textbf{Features} & \textbf{Accuracy} & \textbf{Precision} & \textbf{Recall} & \textbf{F1-score} \\
    \midrule \midrule
        \cite{tao2023androids}~Rand. & Random Classifier & MFCC, RMSE, ZCR, etc. (OpenSMILE) & 50.5 & 55.2 & 55.2 & 55.2 \\
        \cite{tao2023androids}~BS1 & SVM & MFCC, RMSE, ZCR, etc. (OpenSMILE) & \avgstd{73.3}{10.6} & \avgstd{73.5}{16.1} & \avgstd{74.5}{13.2} & \avgstd{73.6}{13.6} \\
        \cite{tao2023androids}~BS2 & LSTM & MFCC, RMSE, ZCR, etc. (OpenSMILE) & \avgstd{83.9}{1.3}  & \avgstd{85.8}{3.1}  & \avgstd{86.1}{2.7}  & \avgstd{84.7}{0.9} \\
        \cite{phukan2024avengers}~FuSeR & PTM & x-vector, emoHuBERT, TRILLsson & 87.9 & -- & -- & 87.8 \\
        \cite{ntalampiras2025interpretable}~UBM-HMM & HMM and MAP & MFCC, TEO, periodicity & \avgstd{87.0}{1.2} & \avgstd{85.8}{1.9} & \avgstd{92.3}{1.8} & \avgstd{88.9}{1.7} \\
        \cite{yu2025using}~PDEM & KELM & PDEM embeddings (based on Wav2Vec2) & \avgstd{88.2}{8.5} & \avgstd{89.6}{6.5} & \avgstd{84.4}{17.9} & \avgstd{86.3}{11.5} \\
        \cite{de2024probing}~HuBERT L & Logistic Regression & HuBERT-L embeddings & -- & -- & -- & 92.0 \\
        \cite{wangCrossLanguageDepressionDetection2025}~MDFA & MDFA\,+GRL & ResNet-18 & -- & -- & -- & 79.2 \\        \cite{wangDepressionDetectionSpeech2026}~MTFS & MTFS-Block & ECAPA-TDNN & -- & 79.4 & 78.8 & 77.4 \\
        \cite{alsenani2024assessing}~Multimodal & GRU & Wav2Vec2 / BERT & \avgstd{86.2}{7.8} & -- & -- & -- \\
        \cite{ilias2024cross}~Concatenation & Cross-Attention & AlexNet (3D log-Mel) / ItalianBERT & \avgstd{91.5}{6.1} & \avgstd{91.5}{8.7} & \avgstd{93.4}{6.0} & \avgstd{92.1}{5.5} \\
        \cite{alsarraniPunctualContinuousAnalyzing2025}~MIL & Majority-voting & OpenSMILE / LIWC & \avgstd{92.5}{1.1} & -- & -- & \avgstd{93.1}{1.1} \\
        \cite{zhangMitigatingInterviewerBias2025}~Full model & D-CoPE\,+\,QF\,+GRL & Wav2Vec2 / XLM-RoBERTa & -- & -- & -- & 73.0 \\
        
    \addlinespace
        ~This work & MultimodalDG & MelSpec / ItalianBERT & \textbf{\avgstd{\accDG}{\accDGstd}} & \textbf{\avgstd{\precDG}{\precDGstd}} & \textbf{\avgstd{\recDG}{\recDGstd}} & \textbf{\avgstd{\FscoreDG}{\FscoreDGstd}} \\
    \bottomrule
\end{tabular}
\end{table*}

\begin{table}[!t]
\caption{Ablation study highlighting the contribution of each modality to model performance.}
\label{tab:ablation_modality}
\centering
\begin{tabular}{ccccc}
    \toprule
    \textbf{Modality} & \textbf{Accuracy} & \textbf{Precision} & \textbf{Recall} & \textbf{F1-Score} \\
    \midrule
    Multimodal  & \textbf{\avgstd{\accDG}{\accDGstd}} & \textbf{\avgstd{\precDG}{\precDGstd}} & \textbf{\avgstd{\recDG}{\recDGstd}} & \textbf{\avgstd{\FscoreDG}{\FscoreDGstd}} \\
    Audio-only  & \avgstd{73.4}{9.5} & \avgstd{74.9}{10.1} & \avgstd{79.0}{10.7} & \avgstd{76.2}{8.3} \\
    Text-only & \avgstd{85.3}{8.1} & \avgstd{87.7}{9.7} & \avgstd{87.7}{9.4} & \avgstd{87.0}{6.0} \\
    \bottomrule
\end{tabular}
\end{table}




\begin{table}[!t]
\caption{Ablation study analyzing the impact of alternative design choices on model performance. Abbreviations include:\\ domain generalization (DG), intra-modal attention (IMA),\\cross-modal attention (CMA), layer normalization (LN).}
\label{tab:ablation_architecture}
\centering
\begin{tabular}{ccccc}
    \toprule
    \textbf{Architecture} & \textbf{Accuracy} & \textbf{Precision} & \textbf{Recall} & \textbf{F1-Score} \\
    \midrule
    Full Model & \textbf{\avgstd{\accDG}{\accDGstd}} & \textbf{\avgstd{\precDG}{\precDGstd}} & \textbf{\avgstd{\recDG}{\recDGstd}} & \textbf{\avgstd{\FscoreDG}{\FscoreDGstd}} \\
    No DG & \avgstd{90.7}{8.9} & \avgstd{90.4}{10.4} & \avgstd{91.9}{7.9} & \avgstd{90.9}{8.5} \\
    No IMA & \avgstd{87.3}{10.4} & \avgstd{85.9}{12.9} & \avgstd{92.6}{6.5} & \avgstd{88.7}{8.9} \\
    No CMA & \avgstd{89.8}{6.8} & \avgstd{89.3}{10.2} & \avgstd{92.5}{6.9} & \avgstd{90.4}{6.4} \\
    No LN & \avgstd{88.9}{7.7} & \avgstd{89.5}{12.1} & \avgstd{92.6}{6.9} & \avgstd{90.2}{6.3} \\
    \bottomrule
\end{tabular}
\end{table}






\subsection{Ablation Studies}
\label{subsec:results_ablation}

A set of ablation experiments was carried out to investigate the contribution of each component in the proposed strategy.

First, the role of each modality was evaluated, as reported in Table~\ref{tab:ablation_modality}. The highest performance was achieved with the full multimodal setup, demonstrating the complementary value of integrating both acoustic and linguistic modalities for accurate depression detection.

The large performance drop in unimodal experiments (e.g., decreasing to 73.4\% accuracy for audio-only) is directly tied to the forced architectural adaptations within $f_\theta$. When restricted to a single modality, the network excludes not only the intra-modally attended BiLSTM stream of the omitted modality (see Fig.~\ref{fig:feature_extraction}), but also inherently disables the CMA module and its subsequent IMA block, as there is no complementary modality to drive the attention computation. Together, these alterations justify the substantial performance gap when compared to the full multimodal framework.

Next, a series of architectural modifications were applied to assess the contribution of each component, as summarized in Table~\ref{tab:ablation_architecture}. The results show that disabling any of these modules inevitably degrades performance, thereby confirming the critical role each plays and underscoring the importance of attention mechanisms, normalization layers, and particularly domain generalization in achieving optimal results.

Notably, the contribution of IMA to the overall performance outweighs that of CMA. IMA highlights salient audio-textual segments to assist subsequent layers in filtering uninformative temporal noise from the BiLSTM and CMA outputs, playing a critical role in achieving optimal performance. In contrast, the architecture is less sensitive to the removal of the cross-modal sequence $\mathbf{seq}_{i,k}^{(c)}$ (see Eqs.~\eqref{formula:seq_cross}--\eqref{eq:latent_features}) from the final latent feature $\mathbf{z}_{i,k}$. This suggests that the subsequent feed-forward network $h_\phi$ can partially compensate for the absence of CMA, likely by relying on the remaining unimodal branches.

\section{Limitations and Future Work}
\label{sec:future}

Despite its methodological contributions and strong results, this study leaves several promising directions for future work. First, due to computational constraints, the modality-specific feature extraction was performed as a stand-alone stage prior to model training. Consequently, the framework is limited by the absence of a fine-tuning stage for the transformer-based encoders, which may reduce performance compared to fully end-to-end approaches. Second, the experiments are limited by the use of a single dataset, due to the scarcity of high-quality, publish-ready datasets in Italian. Future work should explore cross-dataset analysis to confirm the framework's robustness and generalizability, as more datasets become available. Lastly, interpreting internal embeddings under DG settings remains an open challenge, motivating the integration of explainable AI to strengthen clinical trust and enhance the robustness of mental health monitoring systems.

\section{Conclusion}
\label{sec:conclusion}

This study is the first effort to adopt domain generalization for learning domain-invariant features robust to inter-speaker variability in multimodal depression detection. It implements a novel architecture that combines BiLSTM sequence modeling, attention mechanisms, and segment-level decision-making. To identify the most effective audio-text feature extractor pairing, a series of experiments were conducted across various segment durations, ultimately finding the combination of \MEL~and \ITB~with a 30-second duration to be the most effective configuration. Next, this baseline model was integrated with domain-adversarial training, achieving notable improvements in classification performance. The obtained results surpassed all previous benchmarks on the Androids-Corpus, despite the reduced training set size due to validation splitting. Lastly, comprehensive ablation studies confirmed the importance of integrating both modalities and the critical role of each model component. Overall, these findings underscore the applicability of DG as a powerful tool to boost real-world generalizability, enabling patient-independent diagnosis.


    \bibliographystyle{IEEEtran}
    \bibliography{bibliography}
    
\begin{IEEEbiography}[{\includegraphics[width=1in,height=1.25in,clip,keepaspectratio]{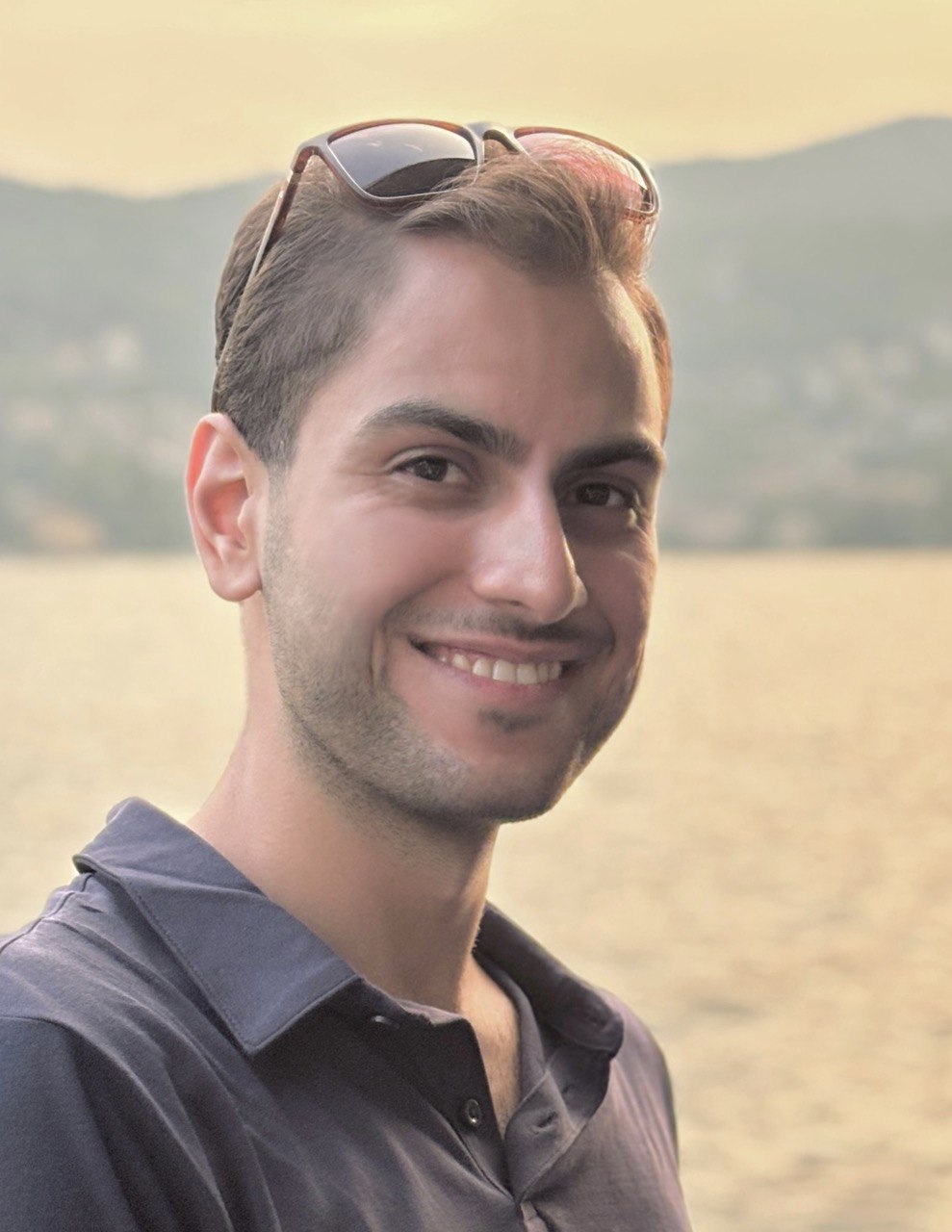}}]{Ali Tabaraei}
is a Ph.D. candidate in Computer Science at the University of Milan, Italy, where he also received his master's degree in the same field in 2025. His current doctoral research focuses on the generalizability and interpretability aspects of advanced deep neural networks applied to health acoustics. Particularly, his research interests include Domain Generalization, Domain Adaptation, Audio Pattern Recognition, Explainable AI, and Health AI.
\end{IEEEbiography}

\begin{IEEEbiography}[{\includegraphics[width=1in,height=1.25in,clip,keepaspectratio]{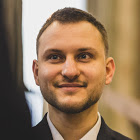}}]{Federico Simonetta}
	is a post-doctoral researcher in the Laudare ERC AdG project at the Gran Sasso Science Institute (GSSI). He previously worked as a post-doctoral researcher at the Universidad Complutense de Madrid and Instituto Complutense de Ciencias Musicales (ICCMU) in the Didone ERC AdG project. He obtained his Ph.D. in Computer Science from the University of Milan in 2022. He is active in the scientific committees of several international journals and conferences. His main research interests are music information processing, machine learning, audio processing, and handwritten music/text recognition.
\end{IEEEbiography}

\begin{IEEEbiography}[{\includegraphics[width=1in,height=1.25in,clip,keepaspectratio]{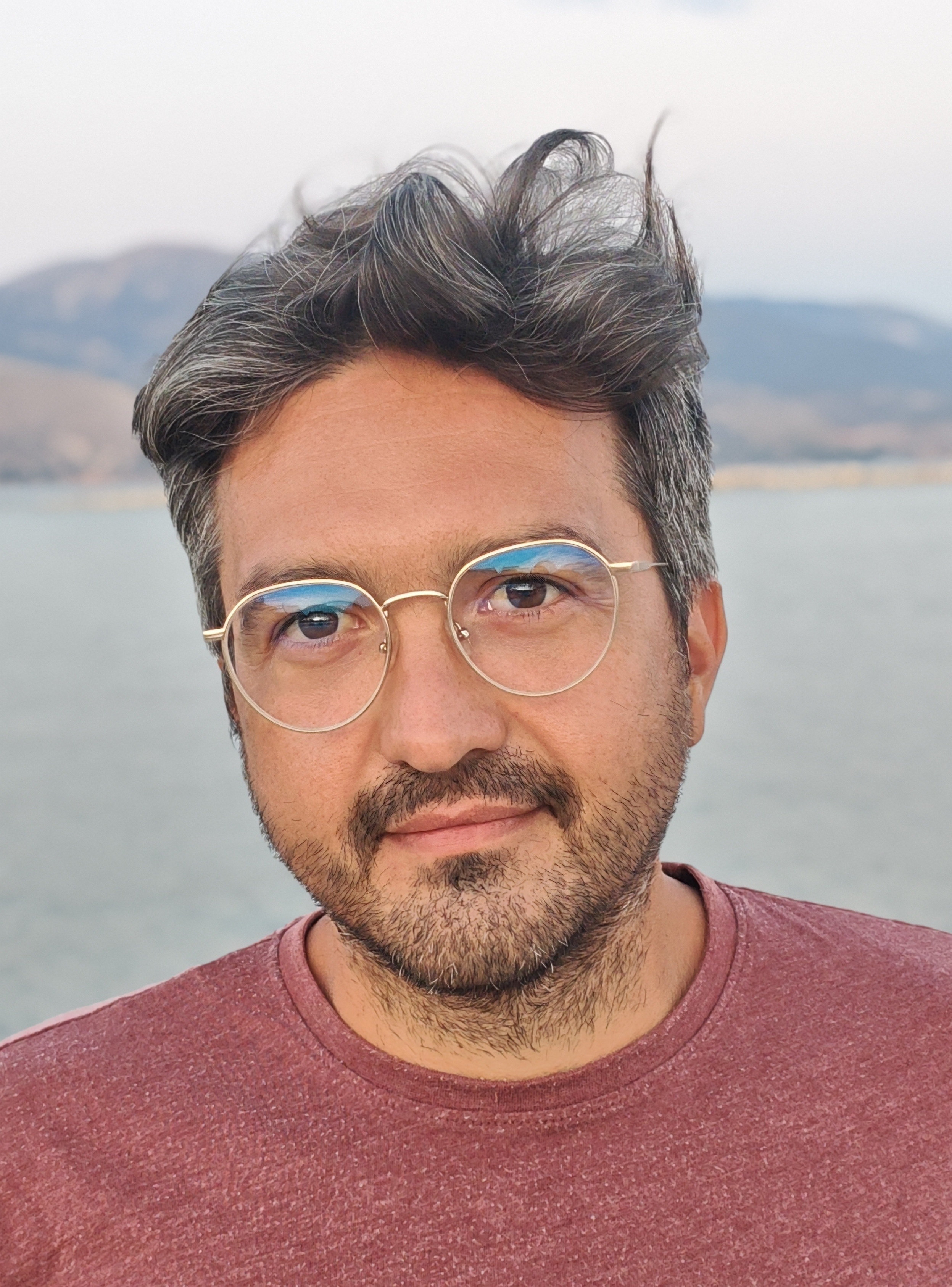}}]{Stavros Ntalampiras} 
 is an Associate Professor at the Department of Computer Science, University of Milan, Italy. He received the engineering and Ph.D. degrees from the Department of Electrical and Computer Engineering, University of Patras, Greece, in 2006 and 2010, respectively. He has carried out research and/or didactic activities at Politecnico di Milano, the Joint Research Center of the European Commission, the National Research Council of Italy, and Bocconi University. Currently, he is an Associate Editor of IEEE TNNLS, PLOS One, IET Signal Processing and CAAI Transactions on Intelligence Technology, as well as member of the IEEE Computational Intelligent Society Task Force on Computational Audio Processing. His research interests include content-based signal processing, machine learning, audio pattern recognition, bioacoustics, health acoustics, and cyber-physical systems.
\end{IEEEbiography}

    \balance

\end{document}